\documentclass[10pt,twocolumn,letterpaper]{article}

\usepackage[pagenumbers]{iccv} 

\usepackage[sort&compress,numbers]{natbib}
\usepackage{mathrsfs}
\usepackage{amsmath}
\usepackage{bm}
\usepackage{multirow}
\usepackage{adjustbox}
\usepackage{graphicx}
\usepackage{colortbl}
\usepackage{microtype} 
\usepackage{amsmath, amssymb, mathtools}
\usepackage{appendix}
\usepackage{wrapfig}
\usepackage{url}
\usepackage{multirow}
\usepackage{caption}
\usepackage{xcolor}
\usepackage{amssymb}
\usepackage{bbm}
\usepackage{graphicx}
\usepackage{amsmath}
\usepackage{enumitem}
\usepackage{booktabs}
\usepackage{adjustbox}
\usepackage{times}
\usepackage{graphicx}      
\usepackage{xcolor}
\usepackage{algorithm}
\usepackage[noend]{algpseudocode}
\usepackage{algorithmicx}
\usepackage{microtype}
\usepackage{lipsum}
\usepackage{graphicx}
\usepackage{subcaption}
\usepackage{caption}
\usepackage{booktabs} %
\usepackage{amsfonts}       %
\usepackage{nicefrac}       %
\usepackage{microtype}      %
\usepackage{comment}
\usepackage{enumitem}
\usepackage{wrapfig,tikz}
\usepackage{bigstrut, tabularx, multirow, makecell, diagbox}
\usepackage{xcolor}

\definecolor{ceruleanblue}{rgb}{0.16, 0.32, 0.75}
\definecolor{iccvblue}{rgb}{0.21,0.49,0.74}
\usepackage[pagebackref,breaklinks,colorlinks,allcolors=iccvblue]{hyperref}

\hypersetup{
colorlinks=true,
citecolor=ceruleanblue,
linkcolor=ceruleanblue}
\usepackage{xspace}

\usepackage[dvipsnames]{xcolor}

\def\paperID{3562} 
\def\confName{ICCV}
\def\confYear{2025}

\title{Moderating the Generalization of Score-based Generative Model}

\author{
Wan Jiang\textsuperscript{1} \quad He Wang\textsuperscript{2} \quad Xin Zhang\textsuperscript{3} \quad Dan Guo\textsuperscript{1} \\ \quad Zhaoxin Fan\textsuperscript{4} \quad Yunfeng Diao$^*$\textsuperscript{1} \quad Richang Hong\textsuperscript{1}\\
\textsuperscript{1}Hefei University of Technology \hspace{1mm}
\textsuperscript{2}University College London \hspace{1mm}\\
\textsuperscript{3}San Diego State University \hspace{1mm}
\textsuperscript{4}Beihang University \hspace{1mm}\\
{\tt\small xjiangw000@gmail.com, he\_wang@ucl.ac.uk, xzhang17@wpi.edu, guodan@hfut.edu.cn} \\{\tt\small zhaoxinf@buaa.edu.cn, diaoyunfeng@hfut.edu.cn, hongrc.hfut@gmail }
}
\begin{document}
\maketitle
\begin{abstract}
Score-based Generative Models (SGMs) have demonstrated remarkable generalization capabilities, \eg generating unseen, but natural data. However, the greater the generalization power, the more likely the unintended generalization, and the more dangerous the abuse. Despite these concerns, research on unlearning SGMs has not been explored. To fill this gap, we first examine the current `gold standard' in Machine Unlearning (MU), \ie, re-training the model after removing the undesirable training data, and find it does not work in SGMs. Further analysis of score functions reveals that the MU ‘gold standard’ does not alter the original score function, which explains its ineffectiveness. Building on this insight, we propose the first Moderated Score-based Generative Model (MSGM), which introduces a novel score adjustment strategy that redirects the score function away from undesirable data during the continuous-time stochastic differential equation process. Albeit designed for SGMs, MSGM is a general and flexible MU framework compatible with diverse diffusion architectures, training strategies and downstream tasks. The code will be shared upon acceptance.
\end{abstract}    
 \section{Introduction}
\label{sec:intro}
``\textit{The greater the power, the more dangerous the abuse.}''
\vspace{-1.5mm}
\begin{flushright}- Edmund Burke \vspace{-2mm}
\end{flushright}
Generative models has been a foundational topic in deep learning in the past decade, \eg Generative Adversarial Networks~\citep{goodfellow2014generative}, Variational Autoencoders~\citep{kingma2013auto} and normalizing flows~\citep{flows}. Recently Score-based Generative models (SGMs)~\citep{song2020score}, Denoising Diffusion Probabilistic Models (DDPMs)~\citep{ho2020denoising} and their variants become one of the dominating class of generative models. In generative models, one key research effort is to maximize their generalization ability, often meaning generating unseen, but natural data, \eg images with realistic faces or stories not written by humans. However, such novel data can also be unintended and potentially cause privacy breaches, copyright infringements, misinformation spreading, \etc. We refer to this phenomenon as \textit{unintended generalization}. Taking SGMs and DDPMs as an example, they can reconstruct training data which should not to be accessible~\citep{carlini2023extracting}, generate faces that are similar to a specific person without permission~\citep{rando2022red,salman2023raising,schramowski2023safe}, and mimic content styles of artists unintentionally~\citep{shan2023glaze,gandikota2023erasing}. Overall, the bigger the generalization power is, the more likely the unintended generalization is, and therefore the more detrimental the intended/unintended harm becomes.
\begin{figure}[!tb]
    \centering     \includegraphics[width=1.0\columnwidth]{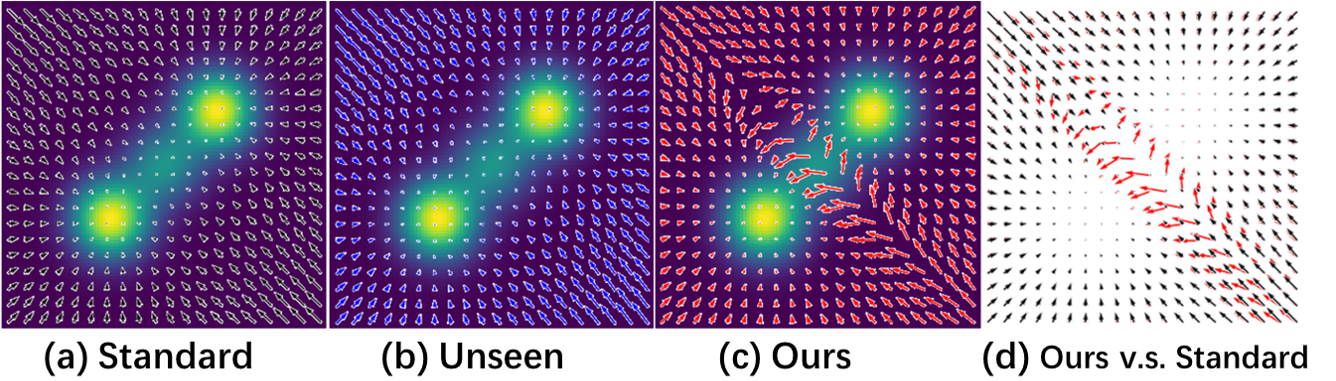}
    \vspace*{-6mm}
    \caption{Comparsions of score functions of Standard VE SDE, Unseen by Re-training and MSGM (Ours) in the toy experiment. Unseen by Re-training retains almost the same score function with Standard VE SDE, while MSGM (red arrows in (d)) significantly alter the original score functions (black arrows in (d)).
    }
    \label{fig:scorepdf}
    \vspace{-6.5mm}
\end{figure}
However, it is not straightforward to design a new Machine Unlearning (MU) paradigm for SGMs and cannot be achieved by adapting existing MU methods for other models. First, MUs for DDPMs primarily aim to reduce the evidence lower bound (ELBO) on the distribution of the forgotten data, but SGMs need to estimate the score function with a continuous noise schedule. This difference in the optimization goals prevents existing DDPM MU methods from being directly applied to SGMs. More generally, most MU methods are designed for conditional generation~\citep{gandikota2023erasing,heng2023continual, fan2024salun, zhang2024forget, kumari2023ablating, wu2024erasediff, heng2024selective,mace}, especially on text-to-image generation with conditional DDPM, tightly coupling the MU with specific conditions, \eg refining cross-attention layers~\cite{mace,gong2024reliable, orgad2023editing,gandikota2024unified,gandikota2023erasing}. They cannot be applied to unconditioned generation, which is widely used for \eg data augmentation. Designing MU for unconditional generation can potentially provide a universal solution to a broader range of generative models. 

To address the challenges, we propose the first Moderated Score-based Generative Model (MSGM) for controlling the unintended generalization. MSGM aims to overcome the limitations of current `gold standard' in score-based generative unlearning, by introducing a straightforward yet effective strategy to alter the score function. Our key idea is to deform the original score function so that it circumvents the Not Suitable For Generation (NSFG) data in sampling, while ensuring that it still approximates the Suitable for Generation (SFG) data score to maintain the generation quality. To this end, we present two variants of MSGM for general MU. The first is to explore the orthogonal complement space of the original score function with respect to the NSFG data, so that sampling will steer away from the high density area of the NSFG data, which works well when the distributions of NSFG and SFG are mildly separable. The second is to explore the negatively correlated score subspace, to target scenarios where the distributions of NSFG and SFG are very similar. Additionally, if NSGF needs to be removed from a pre-trained model, we propose to negate part of the original score function locally. 

Although designed for SGM, MSGM is a general and flexible MU framework compatible with various diffusion models and downstream tasks. To demonstrate its versatility, we conduct extensive experiments on five datasets using SGM, DDPM, and the latent diffusion Model. The results indicate that MSGM consistently achieves strong unlearning performance in both unconditional and conditional generation. Furthermore, it enables zero-shot transfer of unlearning pre-trained models to downstream tasks, including image inpainting and reconstruction.



\section{Related Work}
\label{sec: related_work}

\noindent\textbf{Diffusion Models and Security\&Privacy Issues.} Diffusion models sequentially corrupt training data with slowly increasing noise and then learn to reverse this corruption. DDPMs~\citep{ho2020denoising,nichol2021improved} represent this process as a finite number of denoising steps, while SGMs~\citep{song2019generative,song2020score} generalizes to continuous time using stochastic differential equations (SDEs). Song et al.\citep{song2020improved} improve the training and sampling process and achieve high-fidelity image generation. Meng et al. \citep{meng2021estimating} utilize high-order scores to accelerate the mixing speed of synthetic data and natural image sampling. In addition to technical improvements, SGMs have been shown to be effective across various applications, including natural language processing\citep{pasquale2024generative}, computational physics~\citep{hu2024score}, video prediction~\citep{fiquet2024video}, audio codecs~\citep{wu2024scoredec}, medical imaging~\citep{chung2022score}, \etc. These improvements in diffusion models maximize their generalization ability, but it also raise the concerns about the unintended generalization. Rando et al.~\citep{rando2022red} observed that malicious users may bypass the safety filters in open-source diffusion models to create disturbing content, \eg violence and gore. Beyond this, diffusion models are susceptible to create misleading videos or images of individuals without permission, potentially damaging their reputation or spreading misinformation~\cite{salman2023raising,schramowski2023safe}. Additionally, these models can mimic various art styles, potentially infringing on portrait and intellectual property rights~\citep{shan2023glaze,gandikota2023erasing}.

\noindent\textbf{Machine Unlearning in Generative Model.}
MU enables models to selectively forget the undesirable content for privacy, security or adaptability purposes. Early research has explored numerous MU methods on supervised learning tasks~\citep{shaik2023exploring}, such as image classification~\citep{thudi2022necessity}. However, Fan et al.~\citep{fan2024salun} demonstrated that existing MU for image classification cannot be applied in image generation. This gap highlights the urgent need for effective MU techniques in generative models. Very recently, new MU schemes have been proposed for different types of generative models, including unlearning in VAEs~\citep{moon2024feature,bae2023gradient}, GANs~\citep{kong2023data,sun2023generative} and DDPMs~\cite{ wu2024erasediff}, especially for text-to-image conditional generation~\citep{gandikota2023erasing,heng2023continual, fan2024salun, zhang2024forget, kumari2023ablating, heng2024selective}. However, SGM for MU is still largely missing. More importantly, contrary to the common belief that Unseen by Re-training is the `gold standard' in MU~\citep{thudi2022necessity}, we empirically found it is ineffective in SGM. This motivates us to develop a new SGM MU that can control the unintended generalization and surpass the previous `gold standard' in MU.
\section{Methodology}
\label{sec:method}
\subsection{Preliminaries}
\label{sec:pre}

\noindent\textbf{Machine Unlearning in Generative Model.}
Let $\mathcal{D}= \{\mathbf{x}_i\}^N_{i=1} \in \mathbb{R} ^ D$ be the training data, following the distribution $\mathbf{x}_i \sim p_{d}$. Let $\mathcal{D}_{f} = \{\mathbf{x}^u_i \}^M_{i=1} \subseteq D$ denote the Not Suitable For Generation (NSFG) data following the distribution $p_{f}(\mathbf{x})$. The remaining data, $\mathcal{D}_g = \mathcal{D} \backslash  \mathcal{D}_{f}=\{\mathbf{x}^g_i \}^{N-M}_{i=1} \sim p_{g}(\mathbf{x})$, represents the Suitable For Generation (SFG) data. Our goal is to enable the generative model to avoid generating NSFG samples while maintaining the quality of image generation for SFG data. We refer to such a generative model as an unlearning generative model. We use the symbol $p$ to denote either a probability distribution or its probability density or mass function depending on the context.

\noindent\textbf{Score-Based Generative Modeling with SDEs.}
The two main components of a score-based SDE generative model~\citep{song2020score} are the \textit{forward process} and the \textit{reverse process}. The forward process $\{\mathbf{x}(t) \in \mathbb{R}^d \}_{t=0}^T$ transforming data from the distribution $p_{data}(\mathbf{x})$ to a simple noise distribution with a continuous-time stochastic differential equation (SDE): 
\vspace{-4mm}
\begin{align}
d \mathbf{x} = \mathbf{f}(\mathbf{x}, t) d t + g(t) d \mathbf{w}, t\in [0,T],
\vspace{-3mm}
\end{align}
where $\mathbf{f}:\mathbb{R}^d \to \mathbb{R}^d$ is called the drift, $g \in \mathbb{R}$ is called the diffusion, and $\mathbf{w}$ represents the standard Brownian Motion. Let $p_t(\mathbf{x})$ denote the density of $\mathbf{x}(t)$. At time $t=0$, the initial distribution of $\mathbf{x}(0)$ follows $p_0 := p_{data}$, while at time $t=T$, $\mathbf{x}(T)$ adheres to $p_T$ which is normally an easy-to-sample prior distribution such as Gaussian.
 
Given samples from the prior, the reverse process converts them into data samples via a reverse-time SDE:
\vspace{-1.5mm}
\begin{align}
  d\mathbf{x} = [\mathbf{f}(\mathbf{x}, t) - g^2(t)\nabla_{\mathbf{x}}\log p_t(\mathbf{x})] dt + g(t) d\bar{\mathbf{w}},\vspace{-1.5mm}
\end{align}
where $\bar{\mathbf{w}}$ is a Brownian motion, and $dt$ represents an infinitesimal negative time step. Running the reverse process requires estimating the score function of the forward process, which is typically done by training a neural network with a score-matching objective: 
\vspace{-1.5mm}
\begin{equation}
\label{eq:SGMLOSS}
\begin{aligned}
\min_\theta \mathbb{E}_{t} \lambda(t) \left\{ \right. &\mathbb{E}_{\mathbf{x}(0)} \mathbb{E}_{\mathbf{x}(t)|\mathbf{x}(0)} \left[ \|s_\theta(\mathbf{x}(t), t) \right. \\
& \left.  \left. - \nabla_{\mathbf{x}(t)}\log p_{0t}(\mathbf{x}(t) \mid \mathbf{x}(0))\|_2^2 \right ] \right\}, \\
\end{aligned}
\vspace{-2mm}
\end{equation}
where $\mathbf{x}(0) \sim p_0(\mathbf{x})$ and $\mathbf{x}(t) \sim p_{0t}(\mathbf{x}(t) \mid \mathbf{x}(0))$, $t\sim \mathcal{U}(0,T)$ is a uniform distribution over $[0, T]$, $p_{0t}(\mathbf{x}(t) \mid \mathbf{x}(0))$ denotes the transition probability from $\mathbf{x}(0)$ to $\mathbf{x}(t)$, and $\lambda(t) \in \mathbb{R}_{>0}$ denotes a positive weighting function. Other than~\cref{eq:SGMLOSS}, other score matching objectives, such as sliced score matching \citep{song2020sliced} and finite-difference score matching \citep{pang2020efficient} are also applicable.
\begin{figure*}[!tbh]
    \centering     \includegraphics[width=2.1\columnwidth]{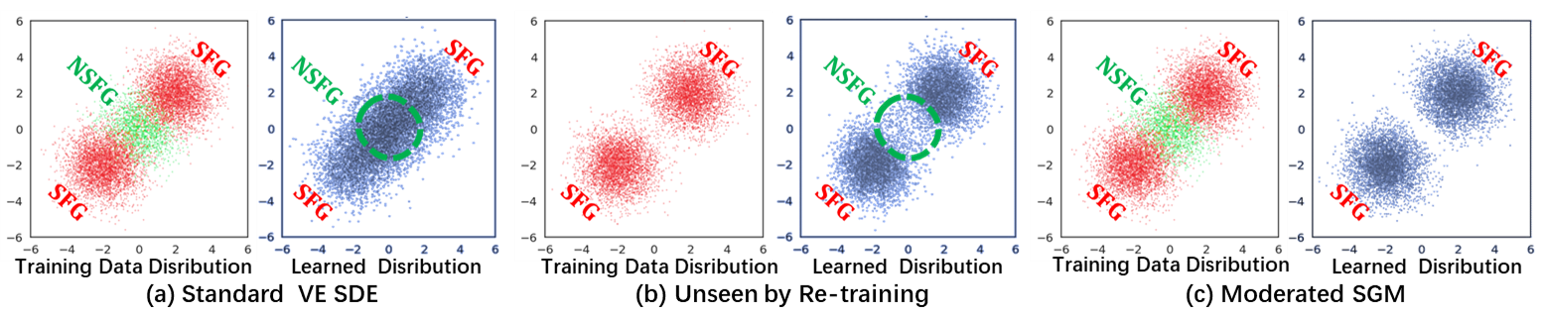}
    \vspace{-8mm}
    \caption{The samples from the mixture Gaussian distribution and the samples generated by the model trained by Standard VE SDE (a), Unseen by Re-training (b) and Unlearning Re-training (c). The left side of (a), (b) and (c) represents the training data, in which the green part is NSFG data, and the red part is SFG data. The right side of (a), (b) and (c) represents the data generated by diffusion models. 
    }
    \label{fig:toysamples}
    \vspace{-5mm}
\end{figure*}
\subsection{Motivation-Unintended Generalization}
\label{sec:motivation}
Two mainstream mechanisms for MU are commonly employed:  (1) \textit{Erasing by Fine-tuning}, which removes learned NSFG features from a pre-trained generator, and (2) \textit{Unseen by Re-training}, which re-trains the generator on filtered data. 
The latter is widely regarded as the `gold standard' in MU~\citep{thudi2022necessity,fan2024salun}, as its superior performance over fine-tuning methods~\citep{xu2024machine}.

Unfortunately, the `gold standard' does not work for SGMs. Specifically, we demonstrate that even after removing NSFG data, SGMs can still generate samples resembling the unlearned features. To formalize this phenomenon, we design a synthetic experiment (\cref{fig:toysamples}) using a mixture of 2D Gaussians:
\vspace{-2mm}
\[
p_{\text{data}} = \textcolor{red}{\underbrace{\frac{4}{5}\mathcal{N}((-2,-2), I)}_{\mathcal{D}_g}} + \textcolor{green}{\underbrace{\frac{2}{5}\mathcal{N}((0,0), I)}_{\mathcal{D}_f}} + \textcolor{red}{\underbrace{\frac{4}{5}\mathcal{N}((2,2), I)}_{\mathcal{D}_g}},\vspace{-2mm}
\]
where $\mathcal{D}_f$ represents the NSFG data and $\mathcal{D}_{g}$ is the rest data. 
The data distribution is shown in \cref{fig:toysamples} (a)\&(c) Left. We train a Variance Exploding Stochastic Differential Equation (VE SDE) model~\citep{song2020score}, referred to as the standard VE SDE, which after training generates a data distribution shown in \cref{fig:toysamples} (a) Right. Clearly, the standard VE SDE learns to generate all data. After using Unseen by Re-training, VE SDE can forget some of the NSFG data, shown in \cref{fig:toysamples} (b) Right, but not completely forget them. This is a typical example of unintended generalization.

\begin{figure}
    \centering     \includegraphics[width=1.0\columnwidth]{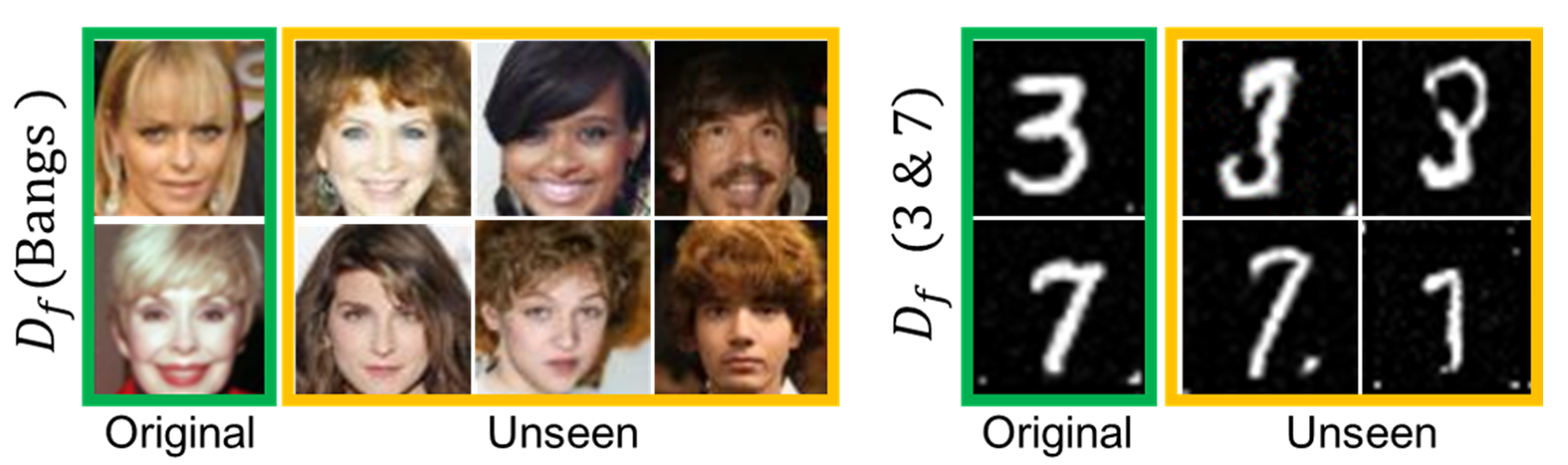}
    \vspace{-7mm}
    \caption{Unintended generalization in high-dimensional data via \mbox{Unseen-by-Re-training}: (Left) CELEBA: Persistent feature retention (yellow boxes: generated ``bangs'') despite exclusion from training data (green boxes). (Right) MNIST: Generation of unlearned classes `3'\&`7' (yellow) omitted from training set (green). 
    }
    \label{fig:ungen}
    \vspace{-5mm}
\end{figure}
Besides visual inspection, we quantify the generation probability of NSFG and SFG data in terms of Negative Log-likelihood (NLL) given different generators in \cref{table:Gauss2NLL}. For Standard VE SDE, it is reasonable for $\mathcal{D}_{g}$ and $\mathcal{D}_{f}$ to have similar likelihoods, as both data are observed during training. However, for Unseen by Re-training, the likelihood of $\mathcal{D}_{f}$ is almost the same as $\mathcal{D}_{g}$. This demonstrates that Unseen by Re-training cannot force the generator to forget the NSFG data well. The visualization of the score functions in \cref{fig:scorepdf} shows that Unseen by Re-training does not alter the original score function of Standard VE SDE, explaining why VE SDE under Unseen by Re-training can still generate NSFG data. 
\begin{table}[!t]
\centering
\caption{The Negative log-likelihood (NLL) values of different methods with respect to the data from $p_\text{data}$.}
\vspace{-2mm}
\label{table:Gauss2NLL}
\begin{adjustbox}{width=0.6\linewidth}
\begin{tabular}{cccc}
\toprule
Test &  Standard&  Unseen& MSGM  \\ \midrule
  $\mathcal{D}_g$   & 10.91  & 10.63  & 10.64\\
  $\mathcal{D}_{f}$  & 10.73  & 11.59   & 39.01\\
\bottomrule
\end{tabular}
\end{adjustbox}
\vspace{-3mm}
\end{table}
We further investigate the existence of unintended generalization in widely-used real-world datasets. Specifically, we conduct experiments on the CELEBA~\citep{alsaafin2017minimal} and MNIST~\citep{liu2015deep} datasets. For the CELEBA dataset, we exclude training samples containing the `bang' attribute. As illustrated in \cref{fig:ungen} (left), the model trained under the Unseen by Re-training paradigm retains the capability to generate realistic images with the `bang' feature, despite its absence from the training data. Similarly, in the MNIST dataset, we remove all samples corresponding to the digits `3' and `7' from the training set. Remarkably, the model still generates digits that closely resemble `3' and `7'. These findings provide empirical evidence that unintended generalization persists in high-dimensional data, challenging the reliability of the current `gold standard' for MU in SGMs. This raises significant concerns regarding the robustness and trustworthiness of existing MU methodologies.

\subsection{Moderated Score-based Generative Model}
To forget the NSFG data, the design object of Unseen by Re-training is to remove $\mathcal{D}_{f}$ from training data and only approximate $p_{g}(\mathbf{x})$, \ie, decreasing the distance between $p_{g}(\mathbf{x})$ and $p_{\theta}(\mathbf{x})$,
\vspace{-2mm}
\begin{equation}
\label{eq:unseenkl}
\theta^{*} = \arg \min_{\theta \in \Theta}  D ( p_\theta (\mathbf{x}), p_{g}(\mathbf{x})),
\vspace{-2mm}
\end{equation}

where the distance between these two distributions $D (p_\theta (\mathbf{x}), p_{g}(\mathbf{x}))$ can be evaluated using some distance metrics. Using score matching, Unseen by Re-training in SGM can train a score network to estimate the score of the distribution $p_{g}(\mathbf{x})$,  
\vspace{-1.5mm}
\begin{align}
\label{eq:sdeg}
L_g& =  \lambda(t) \left\{ \right. \mathbb{E}_{\mathbf{x}(0)}\mathbb{E}_{\mathbf{x}(t)}[\|s^\mathbf{u}_\theta(\mathbf{x}^g(t), t) \\
& -\nabla_{\mathbf{x}^g(t)}\log p_{0t}(\mathbf{x}^g(t) \mid \mathbf{x}^g(0))\|_2^2] \left. \right\}, ~\mathbf{x}^g \in\mathcal{D}_g. \nonumber
\vspace{-1.5mm}
\end{align}
Although Unseen by Re-training has approximated $p_{g}(\mathbf{x})$ and the model generates data that follows $p_{g}(\mathbf{x})$ with high likelihood. However, it does not consider the likelihood of generating $\mathcal{D}_{f}$. If the distributions $p_{f}(\mathbf{x})$ and $p_{g}(\mathbf{x})$ are close or overlapping, Unseen by Re-training may not control the probability of generating $\mathcal{D}_{f}$ (see \cref{fig:toysamples}). Therefore, we propose a new Moderated Generalization strategy to prevent the generator from generating undesired content by maximizing the distance between $p_{f}(\mathbf{x})$ and $p_{\theta}(\mathbf{x})$, while minimizing the distance between $p_{g}(\mathbf{x})$ and $p_{\theta}(\mathbf{x})$, \ie,
\begin{equation}
\label{eq:unlearn}
 \arg \min_{\theta \in \Theta} \left \{D ( p_\theta (\mathbf{x}), p_{g}(\mathbf{x}))  -  D ( p_\theta (\mathbf{x}), p_{f}(\mathbf{x})) \right \}.
\end{equation}

Unlike Unseen by Re-training only approximating $p_{g}(\mathbf{x})$, Moderated Generalization aims to ensure that the generator assigns low likelihood to $\mathcal{D}_{f}$ and high likelihood to $\mathcal{D}_{g}$. However, it is not straightforward to instantiate Moderated Generalization for SGMs. We need to re-consider the Moderated Generalization from the view of score functions. Score estimation is crucial in the generation process of SGMs, as it enables the model to capture data distributions accurately. Theoretically, as long as the score estimation is sufficiently accurate and the forward diffusion time is infinite (allowing the noise distribution to approach the prior distribution), diffusion models can approximate any continuous data distribution with polynomial complexity under weak conditions~\citep{chen2023sampling}. Consequently, \cref{eq:unlearn} can be reframed as a score estimation problem, where different score functions are estimated for $p_{g}(\mathbf{x})$ and $p_{f}(\mathbf{x})$. The challenge then becomes how to train a time-dependent score-based model $s^\mathbf{u}_\theta(\rm \mathbf{x},t)$ to approximate $\nabla_{\mathbf{x}^g} \log p_t(\mathbf{x}^g)$ and deviate $\nabla_{\mathbf{x}^f} \log p_t(\mathbf{x}^f)$. For approximating $p_{g}(\mathbf{x})$, we can directly use \cref{eq:sdeg}. For unlearning $p_{f}(\mathbf{x})$, if the estimated score at any moment deviates from the score of the NSFG data on the timeline from $0$ to $T$, the samples generated during sampling will be far away from the data distribution $p_{f}(\mathbf{x})$. Under this goal, a straightforward idea is to reduce the correlation between $s^\mathbf{u}_\theta(\rm \mathbf{x},t)$ and $\nabla_{\mathbf{x}^f} \log p_t(\mathbf{x}^f)$, i.e. minimizing the dot product of the two scores:
\vspace{-2mm}
\begin{align}
\label{eq:sdeu1}
L_f&  = \lambda(t) \left\{ \right. \mathbb{E}_{\mathbf{x}(0)}\mathbb{E}_{\mathbf{x}(t)}[  \|s^\mathbf{u}_\theta(\mathbf{x}^f(t), t)  \\
&\cdot \nabla_{\mathbf{x}^f(t)}\log p_{0t}(\mathbf{x}^f(t) \mid \mathbf{x}^f(0))\|_2^2]\left.  \right\},  ~\mathbf{x}^f \in\mathcal{D}_f. \nonumber\vspace{-5mm}
\end{align}
\cref{eq:sdeu1} seeks for the orthogonal complement space of $\nabla_{\mathbf{x}^f} \log p_t(\mathbf{x}^f)$, such that for $ \forall \mathbf{x}^f \in \mathcal{D}_{f}$, $s^\mathbf{u}_\theta(\rm \mathbf{x}^f,t) \cdot$   $ \nabla_{\mathbf{x}^f} \log p_t(\mathbf{x}^f) \rightarrow 0 $. We refer to this unlearning optimization as Orthogonal-Moderated Score-based Generative Model, 
or \textit{Orthogonal-MSGM}. However, in our preliminary experiments, we observed that when $p_{g}(\mathbf{x})$ and $p_{f}(\mathbf{x})$ are very close (\eg when generating human faces where local features like bangs or beards are undesirable) or when $s^\mathbf{u}_\theta(\rm \mathbf{x}^f,t)$ has been learned well (\eg erasing undesirable content from a converged pre-trained generator), strictly enforcing the orthogonality constraint in  $\nabla_{\mathbf{x}^f} \log p_t(\mathbf{x}^f)$ becomes challenging. To address this issue, we expand the search space by relaxing the orthogonality constraint to a negatively correlated score subspace, or an Obtuse half-space, defined by $s^\mathbf{u}_\theta(\mathbf{x}^f(t), t) \cdot$ $\nabla_{\mathbf{x}^f} \log p_t(\mathbf{x}^f) < 0$, $ \forall \mathbf{x}^f \in \mathcal{D}_{f}$. This relaxation leads us to propose a new unlearning objective called \textit{Obtuse-MSGM}:
\begin{align}
\label{eq:sdeu2}
L_{f} = & \lambda(t) \left\{ \right. \mathbb{E}_{\mathbf{x}(0) }\mathbb{E}_{\mathbf{x}(t)}[ s^\mathbf{u}_\theta(\mathbf{x}^f(t), t) \\
&\cdot \nabla_{\mathbf{x}^f(t)}\log p_{0t}(\mathbf{x}^f(t) \mid \mathbf{x}^f(0))]\left. \right\},  ~\mathbf{x}^f \in\mathcal{D}_f.\nonumber
\vspace{-2mm}
\end{align}
The final loss of MSGM can be expressed as:
\begin{align}
\label{eq:USGM}
&\min_\theta L_{MSGM} = \min_\theta \mathbb{E}_{t\sim \mathcal{U}(0, T)}   \left ( \alpha L_{g} + 
(1-\alpha) L_{f} \right ), \vspace{-1mm}
\end{align}
where $\mathcal{U}(0,T)$ is a uniform distribution over $[0, T]$, $p_{0t}(\mathbf{x}(t) \mid \mathbf{x}(0))$ denotes the transition probability from $\mathbf{x}(0)$ to $\mathbf{x}(t)$, $\lambda(t) \in \mathbb{R}_{>0}$ denotes a positive weighting function and $\alpha$ is a hyperparameter. 

In contrast to Unseen by Re-training, MSGM modifies the original function of NSFG data. To demonstrate it, we conduct a quick experiment on the mixture Gaussian distribution to evaluate the effectiveness of MSGM. We plot the learned scores at a randomly selected generation process $t=0.08$ in \cref{fig:scorepdf}. The results show that the scores for both Unseen by Re-training and Standard VESDE are quite similar, while our method alters the score distribution of NSFG data, causing the model to steer away from high probability density areas, thereby reducing the likelihood of generating NSFG data.  As shown in \cref{fig:toysamples}, compared to Unseen by Re-training, samples generated by our method almost do not contain NSFG data. Meanwhile, the NLL values in \cref{table:Gauss2NLL} indicate a substantial decrease in the probability of generating NSFG data.

\begin{table*}
\centering
\caption{Quantitative results of unlearning undesirable features or classes using different unlearning methods across various datasets.}
\label{table:mainresult}
\vspace{-0.2cm}
\begin{adjustbox}{width=0.95\linewidth}
\begin{tabular}{cccccccc||cccccc}
\toprule
\multirow{2}{*}{Dataset} & \multirow{2}{*}{Model} & \multirow{2}{*}{Feature/Class} &   \multicolumn{5}{c}{Unlearning Ratio (\%) ($\downarrow$)}   &  \multirow{2}{*}{Test}  & \multicolumn{5}{c}{Negative Log-Likelihood ($\mathcal{D}_g$ ($\downarrow$) and $\mathcal{D}_f$ ($\uparrow$))}  \\ 
\cmidrule{4-8} \cmidrule{10-14} 
&   &  &\multicolumn{1}{c}{Standard} & \multicolumn{1}{c}{Ort} & \multicolumn{1}{c}{Obt} & \multicolumn{1}{c}{Unseen}& \multicolumn{1}{c}{EraseDiff} &  & \multicolumn{1}{c}{Standard} & \multicolumn{1}{c}{Ort} & \multicolumn{1}{c}{Obt} & \multicolumn{1}{c}{Unseen} & \multicolumn{1}{c}{EraseDiff}\\ \cmidrule{1-8} \cmidrule{9-14} 
                    
\multirow{3}{*}{MNIST}   & \multirow{3}{*}{VESDE}  & 3  & 11.0 & \textbf{0.4} & 1.5  & 1.8 &1.4 &   \multirow{2}{*}{ $\mathcal{D}_g$}    & \multirow{2}{*}{2.82}    & \multirow{2}{*}{3.92} & \multirow{2}{*}{3.70}   &\multirow{2}{*}{3.07} &\multirow{2}{*}{3.30}\\

& & 7  &   15.8 & \textbf{0.8} & 3.6   & 2.3  &1.1 &  \multirow{3}{*}{ $\mathcal{D}_f$}   &  &   &  &  \\
& & 3 and 7 &  26.8 & \textbf{1.2}  & 5.1  & 4.1 &2.5 &    &  2.78 & \textbf{13.23}  &12.08 & 3.01 &3.74\\
\midrule
\multirow{3}{*}{CIFAR-10}  & \multirow{3}{*}{VPSDE} &  automobile
&  11.2  &  1.9  & \textbf{0.9}  & 3.4 &9.7 &  \multirow{2}{*}{ $\mathcal{D}_g$}   & \multirow{2}{*}{3.12}  &\multirow{2}{*}{3.22}    & \multirow{2}{*}{3.28}  & \multirow{2}{*}{3.09} &\multirow{2}{*}{3.09}\\
&& dog&  13.4  &  10.0 &  11.5  & 10.8 &\textbf{8.2} &  \multirow{3}{*}{ $\mathcal{D}_f$}  &    &  &  & &\\

&& automobile and dog 
&  24.6  & \textbf{11.9}  &  12.4  & 14.2 &17.9 &   & 3.20  & \textbf{5.94} &4.37 & 3.21 &4.10 \\

\midrule
\multirow{2}{*}{STL-10 }  & \multirow{2}{*}{VPSDE}  &  \multirow{2}{*}{airplane} & \multirow{2}{*}{12.1}& \multirow{2}{*}{\textbf{2.4}} & \multirow{2}{*}{3.6} &  \multirow{2}{*}{3.8}&  \multirow{2}{*}{2.6}& $\mathcal{D}_g$   &  2.90   & 2.90 &  2.92  & 2.90 &-\\
&  &   &  &  &  & &  & $\mathcal{D}_f$   &  2.19  & 8.94 & \textbf{9.25}  & 2.32 &-\\
\midrule
\multirow{1}{*}{CelebA}  & VPSDE & bangs&  19.6 & 3.5 & \textbf{0.7}& 6.7 &1.2 & - &-&- &-&- &-\\
\bottomrule
\end{tabular}
\end{adjustbox}
\vspace{-0.2cm}
\end{table*}
\begin{figure*}[!tbh]
    \centering     \includegraphics[width=1.0\linewidth]{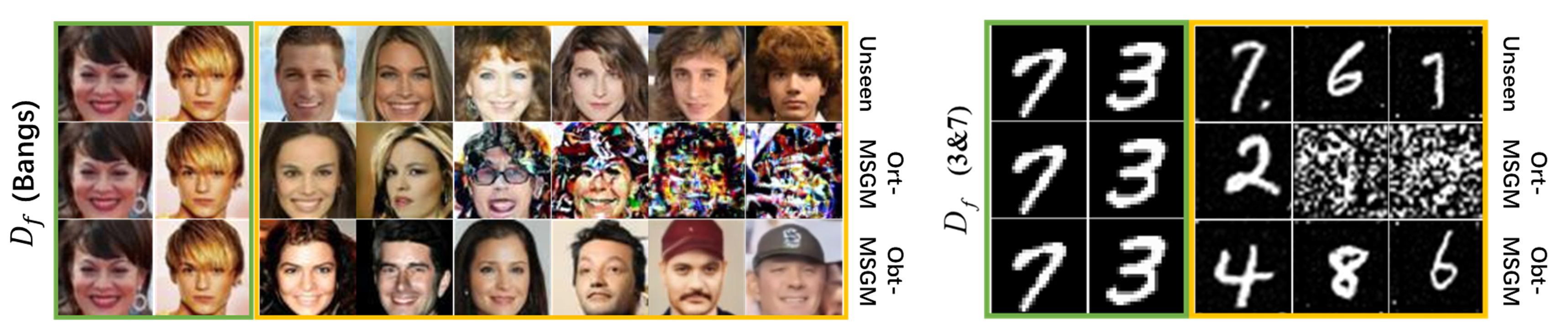}
    \vspace{-6mm}
    \caption{Image generation using different unlearning methods for VE SDE on MNIST and VP SDE on CelebA. The top, middle, and bottom rows show images generated by MU strategy Unseen, Ort and Obt respectively. NSFG images sampled from the forgetting dataset $\mathcal{D}_{f}$ are enclosed in the green box. Images generated by the different unlearning methods are enclosed in the yellow box.
    }
    \label{fig:CEMN}
    \vspace{-0.3cm}
\end{figure*}

\section{Experiments} 
\label{sec:experiments}

\subsection{Experimental Setup}
\textbf{Datasets and Models.} We evaluate MSGM on five datasets, including MNIST~\citep{alsaafin2017minimal}, CIFAR-10~\citep{krizhevsky2009learning}, STL-10~\citep{coates2011analysis}, CelebA~\citep{liu2015deep} and high-resoluation  Imagenette~\citep{howard2020fastai} datasets. In addition to evaluating MSGM on score-based models such as the Variance Preserving (VP) SDE~\citep{song2020score} and the VE SDE~\citep{song2020score}, we also employ DDPM~\citep{ho2020denoising} to verify the generalization of MSGM to different types of diffusion generative models, including the latent diffusion-based Stable Diffusion (SD)~\citep{rombach2022high}. Based on the characteristics of the datasets, we performed class forgetting experiments on MNIST, CIFAR10, STL-10, and Imagenette datasets, while applying attribute elimination generation on CelebA dataset. 

Next, we outline the datasets preparation for the experiments. For MNIST, we trained the VE SDE models, selecting all instances of the digits `3' and `7' for $\mathcal{D}_{f}$. For CIFAR-10, we trained the VP SDE and DDPM models, selecting the data labeled as `dog' and `automobile' classes for $\mathcal{D}_{f}$. For STL-10, we trained the VP SDE models, selecting the data labeled as the `airplane' class for $\mathcal{D}_{f}$. For CelebA, we trained the VP SDE models, selecting the feature `Bangs' from the 40 available features provided for each image to form $\mathcal{D}_{f}$. For Imagenette, we fine-tuned SD v1.4, designating `tench' class as $\mathcal{D}_{f}$.

\noindent\textbf{Compared Methods.}
We establish the following baselines for comprehensive comparison in our experiments: \textbf{Standard}, the conventionally trained generative model without any unlearning intervention, serving as the performance reference; \textbf{Unseen by Re-training (Unseen)}, the gold standard in MU that retrains models from scratch on sanitized datasets; \textbf{EraseDiff}~\cite{wu2024erasediff}, a state-of-the-art diffusion model unlearning method leveraging gradient ascent; and \textbf{ESD}~\cite{gandikota2023erasing}, a parameter-space editing approach for conditional prediction adjustment. 
In our experiments, we evaluate two variants of our proposed method, \textbf{MSGM}: \textbf{Orthogonal-MSGM (Ort)} and \textbf{Obtuse-MSGM (Obt)}. For unconditional generation evaluation, we benchmark both MSGM variants against Standard, EraseDiff, and Unseen baselines. In text-to-image generation tasks, we compare our method against EraseDiff and ESD, as they are specifically designed for conditional generation scenarios. For downstream tasks, we focus on comparing our method against Standard and Unseen, as they represent the two extremes of model behavior—naive training and complete retraining—providing a clear perspective on the trade-offs between performance and unlearning efficacy.

\noindent\textbf{Evaluation Metric.} 
We employ the Unlearning Ratio (UR) and Negative Log-Likelihood (NLL). UR measures the percentage of generated images containing NSFG content, where a lower UR indicates a stronger capability of the model to forget NSFG data. We use external classifiers or CLIP to distinguish whether NSFG categories or features have been removed from the synthetic image. For all experiments, we randomly sample 10,000 images from the model to calculate the unlearning ratio. Additionally, for SGMs, we can accurately calculate NLL to determine the likelihood of generating NSFG and SFG data. Higher values indicate a lower probability of generation. For visual quality evaluation, we follow the protocols of EraseDiff~\citep{wu2024erasediff} to use FID~\citep{heusel2017gans}, CLIP embedding distance~\citep{radford2021learning}, PSNR and SSIM.  


\subsection{Class-wise/Feature-wise Ungeneration}
\textbf{Quantitative Results.} In \cref{table:mainresult}, we compare the unlearning performance with baseline methods in unconditional generation. First, MSGM achieves the lowest unlearning rate compared to Unseen across all datasets, indicating that MSGM effectively unlearns the NSFG data. Second, for Unseen by Re-training, both SFG and NSFG data exhibit low NLL values, suggesting that despite the NSFG data never being observed during the training process, the generative model can still fit the distributions $p_{f}(\mathbf{x})$ well. EraseDiff only slightly decreases the likelihood of generating NSFG data. In contrast, MSGM significantly reduces the generation probability of $\mathcal{D}_{f}$ via substantially increasing the NLL values of the NSFG data. Additionally, although both Orthogonal-MSGM and Obtuse-MSGM can successfully unlearn undesirable data/features, their performance varies across different scenarios. Orthogonal-MSGM is more effective for class unlearning, while Obtuse-MSGM is more effective for feature unlearning. We suspect that Orthogonal-MSGM seeks orthogonal complement space of $\nabla_{\mathbf{x}^f} \log p_t(\mathbf{x}^f)$, so that $s^\mathbf{u}_\theta(\rm \mathbf{x},t)$ does not learn any semantic features(see \cref{fig:CEMN}), hence Orthogonal-MSGM is effective for most cases. However, when $p_{g}(\mathbf{x})$ and $p_{f}(\mathbf{x})$ are very close, the orthogonal complement space of $\nabla_{\mathbf{x}^f} \log p_t(\mathbf{x}^f)$ is hard to be found, hence using Obtuse-MSGM to extend the search space ($s^\mathbf{u}_\theta(\mathbf{x}^f(t), t) \cdot$ $\nabla_{\mathbf{x}^f} \log p_t(\mathbf{x}^f) < 0$, $\forall \mathbf{x}^f \in \mathcal{D}_{f}$) can improve the unlearning performance.

\noindent\textbf{Qualitative Results.} We report the qualitative visualization comparison in \cref{fig:CEMN} and observe that Unseen by Re-training may not fully erase the `bangs' features. For example, facial images generated by Unseen by Re-training may still exhibit few `bangs' features, even though the `bangs' features are not as long as those in $\mathcal{D}_{f}$. In contrast, MSGM completely erases the `bangs' features. An interesting phenomenon is that Orthogonal-MSGM and Obtuse-MSGM forget `bangs' in different ways. For the unwanted feature, Orthogonal-MSGM replaced the `bangs' with noisy images, while Obtuse-MSGM generate features opposite to the `bangs' in the score distribution, such as `no bangs' or `hat'. This occurs because Orthogonal-MSGM seeks the orthogonal complement space of $\nabla_{\mathbf{x}^f} \log p_t(\mathbf{x}^f)$, resulting in $s^\mathbf{u}_\theta(\rm \mathbf{x},t)$ learning nothing. In contrast, $s^\mathbf{u}_\theta(\rm \mathbf{x},t)$ in Obtuse-MSGM learns the inverse of $\nabla_{\mathbf{x}^f} \log p_t(\mathbf{x}^f)$, which may generate the `inverse' feature of `bangs'. The visual results in other datasets also have the similar phenomenon, as shown in the right side of the \cref{fig:CEMN}. Additionally, for SFG content generation, MSGM shows competitive generative performance compared to the source images, and performs well with high-resolution images. 

\subsection{Unlearning DDPM and Fine-tune}
MSGM is a general and flexible framework that is compatible with DDPM models and fine-tuning training. The technical details of MSGM application to DDPM can be found in supplementary material. To demonstrate this, we conduct both class and feature unlearning on pre-trained VP SDE and DDPM models. \cref{table:finetune} and \cref{table:finetuneNLL} present quantitative results for fine-tuning experiments on different datasets using the MSGM method. We conduct 80,000 and 30,000 iterations of fine-tuning on SGM and DDPM architecture respectively, across all datasets. Notably, MSGM not only achieves the best unlearning performance on SGM but also outperforms other baseline on DDPM, even exceeding EraseDiff, an MU method specifically designed for DDPM-based models.

\begin{table}[!t]
\centering
\vspace{-2mm}
\caption{Fine-tune quantitative results for unleaning feature or class on different datasets. The Unlearning Ratio represents the degree of forgetting, measured by predicting the proportion of $\mathcal{D}_{f}$ data in the generated 10,000 images using CLIP. }
\label{table:finetune}
\vspace{-3mm}
\begin{adjustbox}{max width=\linewidth}
\begin{tabular}{cccccccc}
\toprule
\multirow{2}{*}{Dataset} & \multirow{2}{*}{Model} & \multirow{2}{*}{Feature/Class} &   \multicolumn{5}{c}{Unlearning Ratio (\%) ($\downarrow$)}   \\ 
\cmidrule{4-8}
&   &  &\multicolumn{1}{c}{Stand} & \multicolumn{1}{c}{Ort} & \multicolumn{1}{c}{Obt} & \multicolumn{1}{c}{Unseen} & \multicolumn{1}{c}{EraseDiff}\\
\cmidrule{1-8}
\multirow{6}{*}{CIFAR-10}  & \multirow{3}{*}{VPSDE} &  automobile
&  11.2  & 2.7  & \textbf{0.6}  & 3.4  &9.4\\

 &&  dog
&  13.4  &  \textbf{8.7} &  8.9 & 10.8  &5.2\\

&&  automobile and dog
&  24.6  & 11.4  &  \textbf{9.5} & 14.2  &14.6\\
\cmidrule{3-8} 
& \multirow{3}{*}{DDPM} &  automobile
&  13.1  &  3.3  &  \textbf{1.6} & 2.7   & 3.0 \\
 & &  dog
&  13.9  &  5.4 & \textbf{3.6} & 4.5  &4.4 \\
&&  automobile and dog
&  27.0  & 8.7 &  \textbf{5.2} & 7.2  &7.4\\
\midrule
\multirow{1}{*}{CelebA}  & VPSDE & bangs&  19.6 & 2.6 & \textbf{0.1}& 6.7  &1.9 \\
\bottomrule
\end{tabular}
\vspace{-8mm}
\end{adjustbox}
\end{table}
\begin{table}
\centering
\caption{Fine-tuned negative log-likelihood (NLL) values of $\mathcal{D}_{g}$ and $\mathcal{D}_{f}$ for CIFAR-10 data on VP SDE.}
\vspace{-2mm}
\label{table:finetuneNLL}
\begin{adjustbox}{ width=1.0\linewidth}
\begin{tabular}{cccccc}
\toprule
Test &  Standard&  Unseen& Unlearning  &Unseen &EraseDiff\\ \midrule
  $\mathcal{D}_g$   & 2.89  & 3.06  & 4.36 &  2.92 &3.06\\
  $\mathcal{D}_{f}$  & 2.91 & 10.36   & 14.96 & 2.95 &4.38\\
\bottomrule
\end{tabular}
\vspace{-8mm}
\end{adjustbox}
\end{table}
\vspace*{-2mm}

\subsection{Unlearning Text-to-image Generation on High-resolution Datesets}

\begin{figure}[!th]
    \centering     \includegraphics[width=1\linewidth]{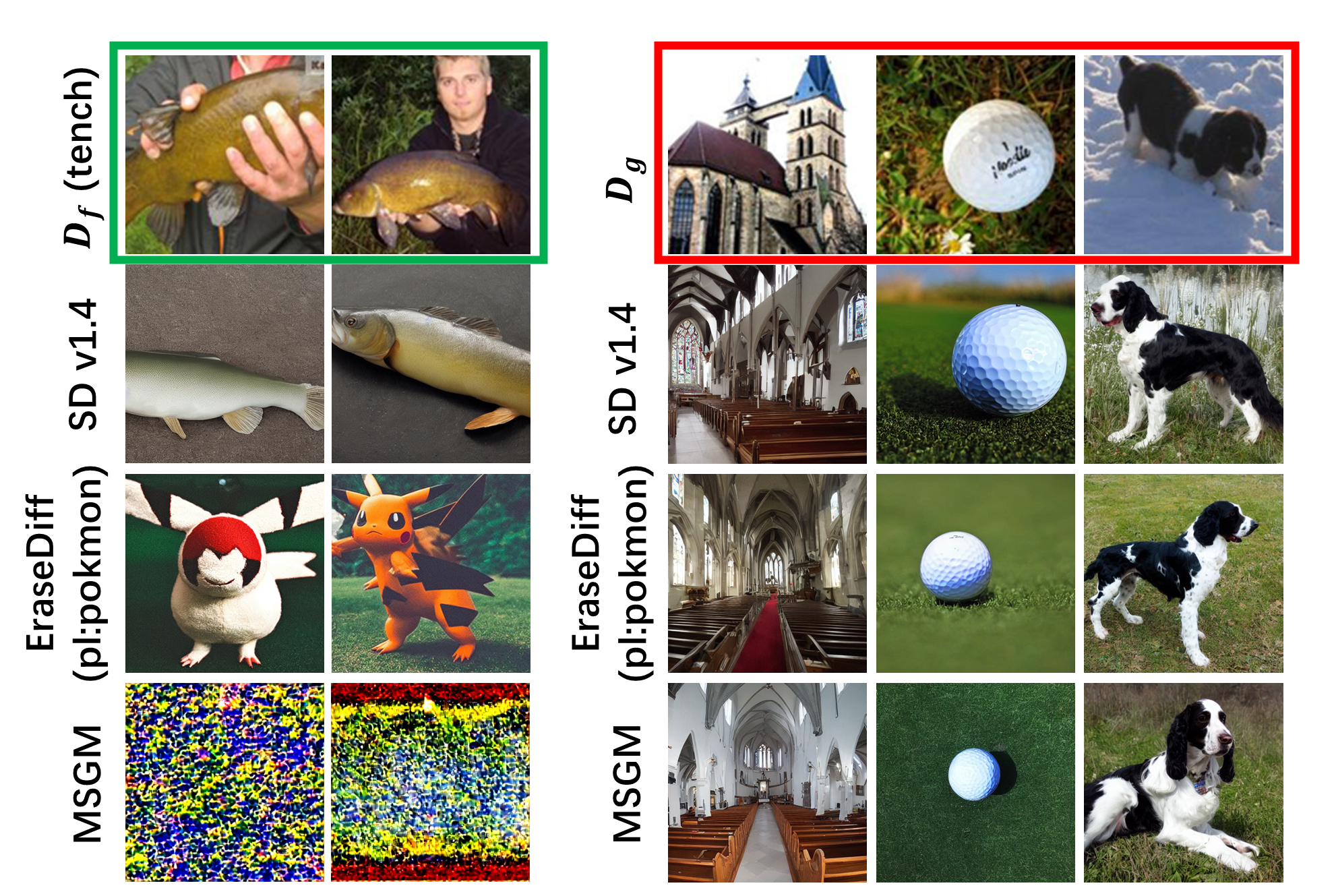}
    \vspace*{-6mm}
    \caption{Visualization of diverse unlearning methods applied to fine-tune SD v1.4 on the Imagenette dataset. The left green box displays NSFG images sampled from forgetting datasets. `pl' indicates the pseudo-label used during training.
    }
    \vspace{-3mm}
    \label{fig:sd}
\end{figure}
\begin{figure}[!tbh]
    \centering     \includegraphics[width=1.0\columnwidth]{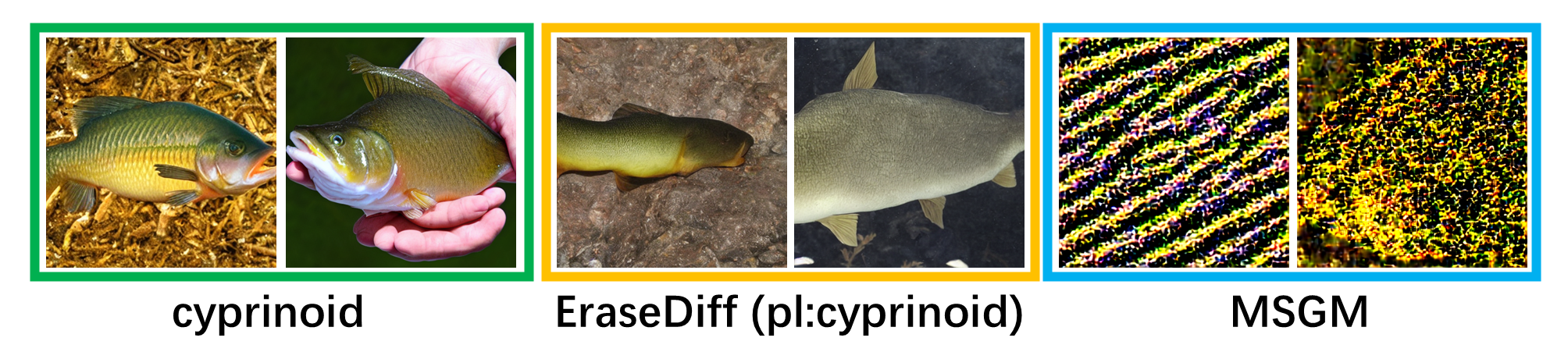}
    \vspace*{-6mm}
    \caption{
Comparison of EraseDiff using semantically similar pseudo-label `cyprinoid' (for `tench') versus our pseudo-label-free MSGM approach.
}
    \label{fig:erasefail}
    \vspace*{-6mm}
\end{figure}
\begin{table}[!t]
\centering
\caption{Unlearning text-to-image models on high-resolution Imagenette dataset.  Classification Rate (CR) measures the classification rate of a pre-trained classifier on generated images conditioned on the forgetting classes.}
\vspace{-3mm}
\label{table:sd}
\begin{adjustbox}{width=0.8\linewidth}
\begin{tabular}{ccccc}
\toprule
 &  SD v1.4 &  ESD & EraseDiff & MSGM \\ \midrule
  FID of $\mathcal{D}_{g}$($\downarrow$)   & 4.89  & 3.09  & 3.09 & 3.08\\ 
  CR($\downarrow$)  & 0.74  & 0.00   & 0.00 & 0.00\\
\bottomrule
\end{tabular}
\end{adjustbox}
\vspace{-5mm}
\end{table}
Although MSGM was initially proposed for unconditional generation, it is a plug-and-play unlearning strategy and also effective for conditional generation, e.g. text-to-image generation. To demonstrate this, we fine-tune Stable Diffusion(SD) v1.4~\citep{rombach2022high} on its cross-attention layer for class-wise forgetting. We compare with other text-to-image models including EraseDiff~\citep{wu2024erasediff} and ESD~\citep{gandikota2023erasing} on high-resolution datatset Imagenette~\citep{howard2020fastai} and choose the `tench' as the forgetting class. As shown in \cref{table:sd}, MSGM achieves an FID of 3.08, outperforming the original SD v1.4 and even slightly improving upon the performance of ESD and EraseDiff. Furthermore, while ESD, EraseDiff, and MSGM all achieve a CR value of $0$, MSGM differs from other methods by directly generating noisy images for the forgotten class, eliminating the need for pseudo-labels from other classes. In contrast, existing methods often rely on pseudo-labels to replace the forgetton class, which can lead to ineffective forgetting when semantically similar categories are used. This limitation frequently results in the failure to fully remove the target class. For instance, both the visualization as shown in \cref{fig:erasefail} and the calculated CR value of 0.47 demonstrate that EraseDiff fails to eliminate the `tench' semantic when the closely related category `cyprinoid' is used as a pseudo-label. MSGM, however, entirely avoids this issue, demonstrating superior robustness and precision in text-to-image unlearning.
\subsection{Application to Downstream Tasks}
\label{sec:downstream}
\begin{figure}[!tbh]
    \centering     \includegraphics[width=1.0\linewidth]{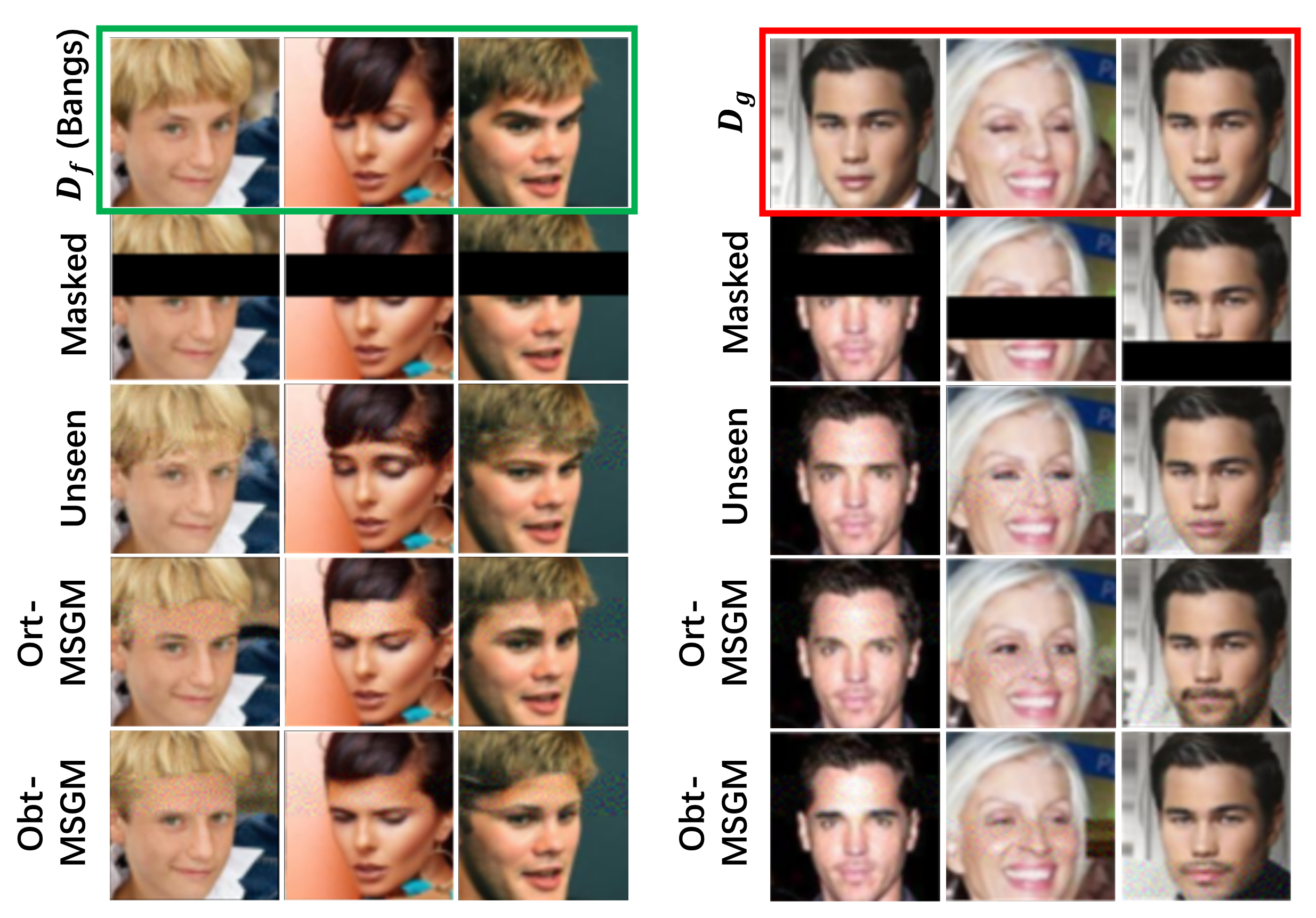}
    \vspace*{-6mm}
    \caption{The comparison of inpainting results on the CelebA dataset. The mask size is $64 \times 16$. The restored results on $\mathcal{D}_{f}$ are displayed on the left. The restored results on $\mathcal{D}_{g}$ are displayed on the right.
    }
    \label{fig:CELEBAIN}
    \vspace*{-5mm}
\end{figure}

\begin{figure}[!tbh]
    \centering     \includegraphics[width=1.0\columnwidth]{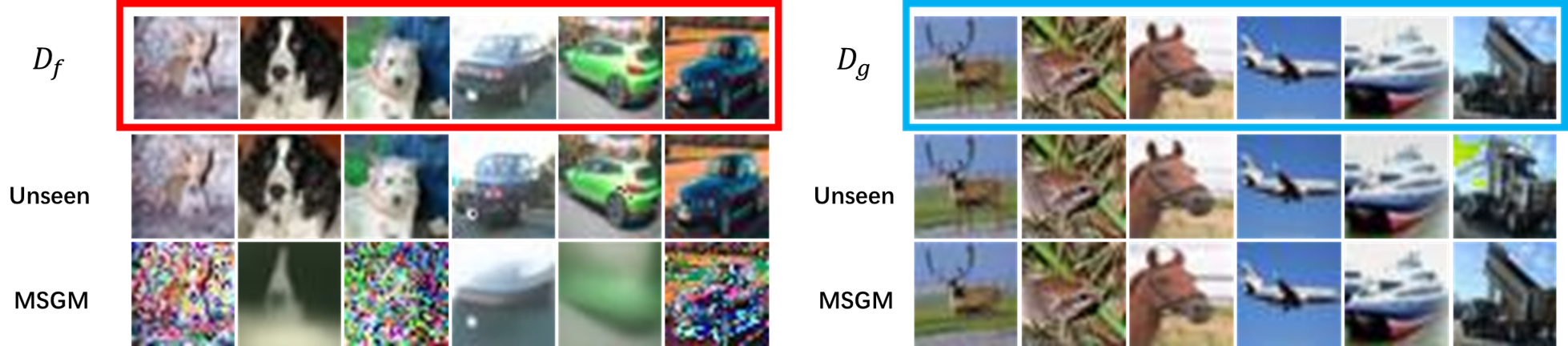}
    \vspace*{-5mm}
    \caption{The comparison of reconstruction results on the CIFAR-10 dataset. The top, middle and bottom columns are the original images, reconstruction images by Unseen, and reconstruction images by Ort respectively. 
}
    \label{fig:recon}
    \vspace*{-4mm}
\end{figure}

\begin{table*}[!tbh]
    \centering
      \caption{The inpainting comparison results.  `Clean' refers to the prediction accuracy on real dataset using the CLIP classifier. `ACC' measures classification accuracy by comparing the predicted class of the inpainted image with the original image's class. The four visual assessment metrics measure the difference between the restored and original images on $\mathcal{D}_{g}$. `stand' refers to the `Standard' method.}
      \label{table:inpaint}
    \begin{adjustbox}{max width=\linewidth}
    \setlength{\tabcolsep}{1.2mm}{
    \begin{tabular}{ccccccc|cccc|cccc|cccc|cccc}
    \hline
    \multirow{2}{*}{Dataset}   & &  \multicolumn{5}{c}{ACC (\%) ($\mathcal{D}_g$($\uparrow$) and $\mathcal{D}_f$($\downarrow$))}  &\multicolumn{4}{c}{FID of $\mathcal{D}_{g}$($\downarrow$)}&   \multicolumn{4}{c}{CLIP of $\mathcal{D}_{g}$($\downarrow$)} &   \multicolumn{4}{c}{PSNR of $\mathcal{D}_{g}$($\uparrow$)} &   \multicolumn{4}{c}{SSIM of $\mathcal{D}_{g}$($\uparrow$)} \\ 
    \cmidrule{3-23} 
    ~ & ~  & Clean & Stand & Ort& Obt& Unseen & Stand & Ort & Obt & Unseen & Stand & Ort& Obt & Unseen& Stand & Ort & Obt & Unseen&Stand & Ort& Obt& Unseen \\
    \midrule
    \multirow{2}{*}{ CIFAR-10}  & $\mathcal{D}_{g}$ & 95.4 & 72.5& 75.5& 74.7 & 75.8 & \multirow{2}{*}{13.11} &\multirow{2}{*}{15.96}&\multirow{2}{*}{13.46}&\multirow{2}{*}{13.64}&\multirow{2}{*}{6.80}&\multirow{2}{*}{6.80}&\multirow{2}{*}{6.77}&\multirow{2}{*}{6.72} & \multirow{2}{*}{31.09}&\multirow{2}{*}{31.09}&\multirow{2}{*}{31.01}&\multirow{2}{*}{31.03} & \multirow{2}{*}{0.56}&\multirow{2}{*}{0.55}&\multirow{2}{*}{0.54}&\multirow{2}{*}{0.54} \\ 
        ~ &     $\mathcal{D}_{f}$  & 95.5 & 75.0 & 57.2 & \textbf{49.6} & 59.7 &&&& &&&& &&&& \\ \hline
         \multirow{2}{*}{ STL-10}   &$\mathcal{D}_{g}$ & 96.3 & 83.4 & 83.6 &83.1 & 84.5& \multirow{2}{*}{28.48} &\multirow{2}{*}{29.95}&\multirow{2}{*}{28.55}&\multirow{2}{*}{28.56}& \multirow{2}{*}{8.50} & \multirow{2}{*}{8.51} &  \multirow{2}{*}{8.50}& \multirow{2}{*}{8.50} & \multirow{2}{*}{31.18}&\multirow{2}{*}{31.17}& \multirow{2}{*}{31.17}& \multirow{2}{*}{31.18}&\multirow{2}{*}{0.59}&\multirow{2}{*}{0.58}&\multirow{2}{*}{0.57}&\multirow{2}{*}{0.59}\\ 
        ~ &  $\mathcal{D}_{f}$ & 96.3 & 84.1& 59.5 & \textbf{50.3} & 54.9 &  &  & &  &  &  & &&&&&\\ \hline
         \multirow{2}{*}{ CelebA} & $\mathcal{D}_{g}$ & 98.3 &95.5 & 99.0& 99.5 & 98.0 &\multirow{2}{*}{29.42}&\multirow{2}{*}{30.31}&\multirow{2}{*}{29.43}&\multirow{2}{*}{30.42}& \multirow{2}{*}{8.96} &\multirow{2}{*}{ 8.96} &\multirow{2}{*}{8.97} &\multirow{2}{*}{8.94} & \multirow{2}{*}{34.54} &  \multirow{2}{*}{34.52}&\multirow{2}{*}{34.50} & \multirow{2}{*}{34.54} &\multirow{2}{*}{0.83}& \multirow{2}{*}{0.82} & \multirow{2}{*}{0.81} & \multirow{2}{*}{0.82} \\
        ~ &   $\mathcal{D}_{f}$ & 98.3  &53.0& 1.0 & \textbf{0.5} & 2.0&  & &   &  &  & &  & &&&&\\\hline
    \end{tabular}}
    \end{adjustbox}
    \vspace{-0.2cm}
\end{table*}

\begin{table*}[!tbh]
    \centering
      \caption{The comparison results of reconstruction, where `stand' denotes the `Standard' method.}
      \label{table:recon}
    \vspace{-0.2cm}
    \begin{adjustbox}{max width=\linewidth}
    \setlength{\tabcolsep}{1.2mm}{
    \begin{tabular}{ccccccc|cccc|cccc|cccc|cccc}
    \hline
    \multirow{2}{*}{Dataset}   & &  \multicolumn{5}{c}{ACC (\%) ($\mathcal{D}_g$($\uparrow$) and $\mathcal{D}_f$($\downarrow$))}  &   \multicolumn{4}{c}{FID of $\mathcal{D}_{g}$($\downarrow$)} &
    \multicolumn{4}{c}{CLIP of $\mathcal{D}_{g}$($\downarrow$)} &
    \multicolumn{4}{c}{PSNR of $\mathcal{D}_{g}$($\uparrow$)} &   \multicolumn{4}{c}{SSIM of $\mathcal{D}_{g}$($\uparrow$)} \\ 
    \cmidrule{3-23} 
    ~ & ~  & Clean & Stand & Ort & Obt & Unseen & Stand & Ort & Obt & Unseen & Stand & Ort & Obt & Unseen& Stand & Ort& Obt & Unseen& Stand & Ort& Obt & Unseen\\
    \midrule
    \multirow{2}{*}{ CIFAR-10}  & $\mathcal{D}_{g}$ & 95.4 & 88.1 &87.7  &87.0 & 87.9& \multirow{2}{*}{5.52}&  \multirow{2}{*}{5.71 } &\multirow{2}{*}{5.57}&\multirow{2}{*}{5.94}& \multirow{2}{*}{6.91}&\multirow{2}{*}{6.90}&\multirow{2}{*}{6.89}&\multirow{2}{*}{6.90} & \multirow{2}{*}{ 31.91}&\multirow{2}{*}{32.15}&\multirow{2}{*}{32.19}&\multirow{2}{*}{31.82} & \multirow{2}{*}{0.92}&\multirow{2}{*}{0.92}&\multirow{2}{*}{0.93}&\multirow{2}{*}{0.91} \\ 
        ~ &     $\mathcal{D}_{f}$  & 95.5 & 74.4 & \textbf{48.4} & 69.6 & 70.3 &&&& &&&& &&&& \\ \hline
        
    \end{tabular}}
    \end{adjustbox}
    \vspace{-4mm}
\end{table*}

\noindent \textbf{Unleanring Inpainting.}
MSGM enables zero-shot transfer of the SGM MU to downstream tasks. We first test MSGM on inpainting task. For the class inpainting on CIFAR-10 and STL-10, we mask the upper half of the image and attempt to restore the whole image. For feature inpainting, we mask the region of the feature to be restored (covering 1/4 of the whole image, shown in \cref{fig:CELEBAIN}). The quantitative restoring results on $\mathcal{D}_{f}$ and $\mathcal{D}_{g}$ are reported in \cref{table:inpaint}. We regard the classification as correct if the predicted class of the restored image matches that of the corresponding original image. Obtuse-MSGM still contains a high classification accuracy for restored images on $\mathcal{D}_{g}$ while significantly decrease the accuracy on restored images on $\mathcal{D}_{f}$. This indicates that restored image by Obtuse-MSGM still retains similar semantics on $\mathcal{D}_{g}$, while altering the source semantics on $\mathcal{D}_{g}$. 

Next, For the image quality metrics (FID, CLIP, PSNR and SSIM) on $\mathcal{D}_{g}$, there is no significant difference between MSGM and the standard trained model, indicating that the MSGM still retains high visual quality. Furthermore, we compare the visual results on \cref{fig:CELEBAIN}. When the masked image is from $\mathcal{D}_{f}$, Unseen by Re-training still has the probability to restore the `bangs' feature in the masked region, while MSGM effectively erase the bangs on $\mathcal{D}_{f}$. When the masked image is from $\mathcal{D}_{g}$, MSGM can successfully restore realistic masked features, such as forehead, nose, mouth \etc.

\noindent\textbf{Unlearning Reconstruction.} Generative models can learn the latent representations of data and reconstruct images. Through the reconstruction, we use these latent representations as guidance to verify whether our method effectively achieves unlearning. To maintain the similarity between reconstruction results and original images on $\mathcal{D}_{g}$, we set $t = 0.02$ for the continuous-time SDE schedule. We reconstruct images using VP SDE model trained by standard training, Unseen by Re-training and MSGM, and report the comparison results in \cref{table:recon}. We utilize the classification accuracy to assess whether the reconstructed images still be classed by the original class. Obtuse-MSGM significantly decrease the accuracy for reconstructed $\mathcal{D}_{f}$ data while maintaining the original semantic information for reconstructed $\mathcal{D}_{g}$. Additionally, we calculate the CLIP distance, PSNR and SSIM between original images and reconstructed images on $\mathcal{D}_{g}$. The images reconstructed by MSGM and the standard model have nearly the same numerical results, indicating that the images reconstructed by MSGM have high visual quality. Next, we visualize the reconstruction results in \cref{fig:recon}. Unlike Unseen by Re-training, where the reconstruction images on $\mathcal{D}_{f}$ have the same semantics with the original images, Orthogonal-MSGM reconstructs $\mathcal{D}_{f}$ as noisy images, indicating that Orthogonal-MSGM has completely unlearned the $\mathcal{D}_{f}$ distribution. 

\begin{table}[!t]
\centering
\caption{Ablation studies on MNIST with Orthogonal-MSGM using different parameters $\alpha$. $\alpha=1$ means Unseen.}
\vspace{-2mm}
\label{table:alpha}
\begin{adjustbox}{ width=1\linewidth}
\setlength{\tabcolsep}{1.3mm}{
\begin{tabular}{ccc|cc||ccc|cc}
\toprule
$\alpha$& Class &    UR(\%) ($\downarrow$)  &  \multicolumn{2}{c||}{NLL Test } & $\alpha$& Class &    UR(\%) ($\downarrow$)  &  \multicolumn{2}{c}{NLL Test } \\ 
\midrule
\multirow{3}{*}{0.7}    & 3  & 0.7  &   \multirow{2}{*}{ $\mathcal{D}_g$}    & \multirow{2}{*}{4.17}  &\multirow{3}{*}{0.99}    & 3  & 0.4 &   \multirow{2}{*}{ $\mathcal{D}_g$}    & \multirow{2}{*}{3.92}  \\
 &  7  &   1.2 &   \multirow{3}{*}{ $\mathcal{D}_f$}   &   &&  7  &   0.8 &   \multirow{3}{*}{ $\mathcal{D}_f$}   &   \\
 & 3 and 7 &  1.9 & &13.32 && 3 and 7 &  1.2 & &14.75 \\
\midrule
\multirow{3}{*}{0.9}    & 3  & 0.5  &   \multirow{2}{*}{ $\mathcal{D}_g$}    & \multirow{2}{*}{4.05}  &\multirow{3}{*}{1.0}    & 3  & 1.8 &   \multirow{2}{*}{ $\mathcal{D}_g$}    & \multirow{2}{*}{3.07} \\
&  7  &   0.9 &   \multirow{3}{*}{ $\mathcal{D}_f$}   &  &&  7  &   2.3 &   \multirow{3}{*}{ $\mathcal{D}_f$}   & \\
& 3 and 7 &  1.4& &13.88  && 3 and 7 & 4.1 & &3.01\\
\bottomrule
\end{tabular}}
\end{adjustbox}
\vspace{-5mm}
\end{table}

\subsection{Ablation Study}
\label{sec:ablation}
\begin{figure}[!t]
    \centering     \includegraphics[width=1.0\linewidth]{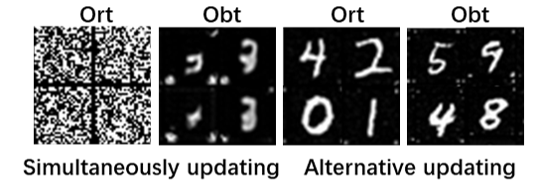}
    \vspace*{-8mm}
    \caption{The visual results of Simultaneously Updating vs. Alternative Updating. Unlike Alternative Updating generating realistic images, Simultaneously Updating generate noisy or blur images.
    }
    \label{fig:iter}
    \vspace*{-5mm}
\end{figure}

\noindent\textbf{Optimization Choices.} We explore two strategies for optimizing \cref{eq:USGM}: (1) Simultaneously Updating: both $L_{g}$ and $L_{f}$ are simultaneously updated in each iterative sampling, and (2) Alternative Updating: $L_{g}$ is updated in each iterative sampling, while $L_{f}$ is updated at intervals of every four iterations. We conduct experiments on MNIST using VE SDE, with $\alpha$ set to 0.7. We plot the loss curve in supplementary material and visualize the visual results in \cref{fig:iter}. We find that $L_{g}$ converges much more easily than $L_{f}$. As a result, the alternative updating help $L_{g}$ converge more effectively and can improve visual quality. Consequently, we adopt the alternative updating strategy.

\noindent \textbf{$\alpha$ Setting.} There are two key loss functions in MSGM(\cref{eq:USGM}), in which $\alpha L_{g}$ ensures the original generating quality while $(1-\alpha) L_{f}$ avoids to generate NSFG data. Therefore, we conduct an ablation study to evaluate their impacts in \cref{table:alpha}. We found that when $\alpha$ is set to 0.99, our method achieves a good balance between the generation quality and unlearning performance. We hence set $\alpha = 0.99$ as default.
\section{Conclusion and Future work}
In this work, we make the first attempt to investigate SGM MU and propose the MSGM, a novel framework that surpasses the current `gold standard' and existing DDPM-based unlearning approaches. Extensive experiments show that MSGM effectively unlearns undesirable content while preserving the generation quality of retain data. Albeit primarily designed for SGMs, MSGM is a plug-and-play unlearning strategy applicable to diverse diffusion architectures and training methods. It also enables zero-shot transfer of the pre-trained models to downstream tasks, showcasing robustness in unlearning under inappropriate content guidance. Future work will extend MSGM to more data types, such as time-series skeletal motion and video data.
{
    \small
    \bibliographystyle{ieeenat_fullname}
    \bibliography{main}
}
\end{document}